\documentclass[cameraready]{Interspeech}

\title{The Lipreading Gap: Do VSR Models Perceive Visual Speech Like Human Lipreaders?}

\author[affiliation={1}, orcid=0000-0002-4891-494X]{Rishabh}{Jain}
\author[affiliation={1}, orcid=0000-0002-9274-209X]{Naomi}{Harte}

\address{
    $^1$ Sigmedia Group, School of Engineering, Trinity College Dublin, Ireland
}

\email{rijain@tcd.ie, nharte@tcd.ie}

\keywords{visual speech recognition, lipreading, human perception, human-machine alignment, viseme analysis}

\usepackage{comment}
\usepackage{booktabs}
\usepackage[table]{xcolor}  
\usepackage[utf8]{inputenc}
\usepackage{textgreek} 
\usepackage[utf8]{inputenc} 
\usepackage{tipa}           
\usepackage{makecell}
\usepackage{bm}

\begin{document}

\maketitle

\begin{abstract}

Visual speech recognition (VSR) models now surpass human lipreaders on benchmarks, but do such gains establish human-like visual speech perception? To explore this, we compare three VSR systems with human baselines on the MaFI word-level lipreading dataset using word, character, phoneme, and viseme-level metrics. Although models achieve higher overall accuracy, they succeed and fail on different words than humans. A text-only n-gram baseline given only a few initial phonemes rivals human lipreading. VSR word-level errors are consistently better explained by training word frequency than by the visual informativeness of words. Viseme accuracies, confusion matrices and human–model correlations further show that models gain most on visemes humans find hardest, and show much weaker dependence on visual clarity.  Our work demonstrates that VSR systems rely primarily on language cues from training data rather than visual perception, failing to bind visual features into meaningful words.

\end{abstract}

\section{Introduction}

Visual Speech Recognition (VSR) has achieved remarkable performance on standard benchmarks. Modern transformer-based models achieve word error rates (WER) below 17\% on LRS3 \cite{liu2023synthvsr,son2017lip}, representing a substantial improvement over earlier approaches. These advances in architecture and self-supervised pretraining suggest that machines have mastered lipreading \cite{prajwal24_interspeech, ahn24_interspeech}. However, high benchmark accuracy does not guarantee that VSR models process visual speech using the same mechanisms as humans \cite{geirhos2020beyond,lin2025uncovering, bowers2023deep}. Recent work shows that VSR models trained on LRS3 \cite{lrs3} often fail to generalize to new, unseen datasets \cite{djilali2024vsr, jain2026hypeinsightrethinkinglarge}, suggesting over-reliance on training-specific patterns. While VSR models are trained on continuous speech where sentence-level context is available, it remains unclear whether their recognition relies on visual articulatory cues or learned sequential patterns.  While humans rely on visual articulation cues to perceive speech \cite{battista2025lip, bernstein2014neural}, whether machines exploit similar visual information remains unclear \cite{lin2025uncovering, lippmann1996speech}. This raises a fundamental question: do VSR models achieve high accuracy through visually-grounded speech processing or through learned linguistic patterns that bypass direct visual understanding \cite{Hong_2023_CVPR,lin2025uncovering}? 

Answering this question is challenging because transcription accuracy alone cannot reveal underlying perceptual mechanisms. A model could succeed by exploiting learned linguistic context or co-occurrence patterns rather than bottom-up articulatory features \cite{karas2019visual}. Moreover, comparative analyses between humans and VSR systems remain limited \cite{lopez, hilder2009comparison, potamianos2001large, benoit1996components}. Early work indicated differences in word versus phoneme-level performance between humans and machines \cite{hilder2009comparison}, but these studies relied on small datasets and traditional statistical models such as Hidden Markov Models (HMM)  \cite{hilder2009comparison,kong2017evaluating, 6288999, chaloupka04_interspeech, karpov10_interspeech, tao14_interspeech, noda14_interspeech}. No prior research has systematically compared human and state-of-the-art transformer-based models directly. Although viseme patterns have been studied in machine lip-reading \cite{benoit1996components,7472029, Hong_2023_CVPR,BEAR201740, hu-etal-2023-hearing, kashiwagi23_interspeech}, whether VSR models exhibit human-like viseme confusions remains unknown. Likewise, there has been no exploration of whether models respond to visual clarity in the same manner as humans \cite{MAVSR}. In related computer vision domains, research shows that alignment between human perceptual behavior and model predictions strongly predicts generalization and robustness \cite{muttenthaler2022human, lee2023visalign, afshan22b_interspeech, mansfield21_interspeech}. Despite these findings, VSR evaluation continues to rely primarily on transcription metrics such as WER, which overlook perceptual aspects of visual speech understanding \cite{lin2025uncovering, advocating_cer, avsuperb}.

A fundamental barrier to such work has been the absence of publicly-available human lipreading data. The MaFI benchmark \cite{krason2024mouth} addresses this by providing human lipreading accuracy and visual informativeness scores for English words. Comparing performance on isolated words could disentangle reliance on visual articulatory cues from reliance on learned sequential patterns. This design allows us to evaluate whether models develop visual understanding that generalizes beyond their continuous speech training data, or whether they rely on learned sequential patterns that fail when sentence-level context is unavailable. We leverage the MaFI dataset to conduct the first detailed quantitative analysis of human-machine alignment \cite{millet-dunbar-2022-self} in VSR. We compare three state-of-the-art models: Auto-AVSR \cite{autoavsr} (supervised), AV-HuBERT \cite{avhubert} (self-supervised), and VSP-LLM \cite{vspllm} (LLM-based), against human baselines using multi-level, multi-metric correlation analysis, viseme-level comparisons \cite{BEAR201740}, and confusion pattern examination across visual clarity levels. Our analysis reveals systematic divergences between human and models' visual speech processing. Text-only baselines achieve substantial accuracy without visual input, and model errors correlate more with training frequency than visual difficulty. Humans show strong correlation between visual clarity and recognition accuracy, while VSR models show substantially weaker correlations, demonstrating fundamentally different visual speech processing mechanisms. 

Our paper makes multiple valuable contributions. First, we establish recognition baselines for humans versus VSR models using multi-level metrics (word, character, phoneme, and viseme). Second, we quantify the role of language patterns independent of visual processing through text-only n-gram baselines and training frequency analysis. Third, we provide viseme-level analysis comparing human and VSR model performance using consistent phoneme-to-viseme mapping, highlighting differences in accuracy and confusion patterns. Fourth, we quantify human-model alignment through correlation analysis to assess whether models succeed on the same words and make similar errors. Fifth, we examine how models and humans respond to visual clarity through performance analysis and correlation metrics across clear and ambiguous words. We reveal systematic differences in visual speech perception between humans and models.

\section{Materials and Methods}
\label{sec:method}

\subsection{Dataset Description}

We conduct our analysis using the MaFI dataset \cite{krason2024mouth}, which provides Mouth and Facial Informativeness (MaFI) scores for 2,276 English words based on human lipreading experiments conducted with 410 participants (263 native British English speakers and 147 native North American English speakers). In these experiments, participants watched silent video clips of speakers articulating words and attempted to identify them through lipreading alone, based solely on visible mouth and facial movements. The dataset spans diverse lexical categories and phonetic structures, providing comprehensive coverage of visual speech information across an English vocabulary. For our evaluation, we select the subset of MaFI words where corresponding video data is available for all words attempted by human participants during the lipreading experiments. This results in 2,189 word-level test samples out of the original 2,276 words in the MaFI dataset. 

\subsection{MaFI Score}
The MaFI score developed in \cite{krason2024mouth} measures how visually informative words are, based on their corresponding mouth and facial movements. These scores are highly consistent across perceivers of different English accents and are quantified using the phonological distance between participants' lipreading responses and target words. Each word was independently rated by at least 10 participants. Scores range from approximately -2.5 to 0, with values near 0 indicating higher visual informativeness (words that are easier to lipread) and values near -2.5 indicating lower informativeness (visually ambiguous words). MaFI scores are computed using the Levenshtein distance \cite{Levenshtein1965BinaryCC} between participant responses and target words, capturing phonetic and visual similarity rather than binary correctness. This provides a more nuanced measure of visual speech information that reflects perceptual similarity in human recognition.

\subsection{Pretrained VSR Models}

\begin{table}[t]
  \caption{VSR models and their performance on LRS3.}
  \label{tab:models}
  \centering
  \footnotesize
  \begin{tabular}{|l|c|c|c|c|}
    \hline
    \rowcolor{green!30!blue!20}
    \textbf{Model} & \textbf{Type} & \textbf{Data (Hours)} & \textbf{Size} & \textbf{LRS3 \bm{$\downarrow$}} \\
    \hline
    Auto-S & SL & 1,759h labeled & 250M & 24.6\%\\
    \hline
    Auto-L & SL & 3,291h labeled & 250M & 20.3\% \\
    \hline
    AV-HuBERT & SSL &PT/FT: 1,759h/433h& 95M & 28.6\% \\
    \hline
    VSP-LLM & LLM &PT/FT: 1,759h/433h& 7B & 25.4\% \\
    \hline
  \end{tabular}
  \\[2pt]
  \footnotesize{\textbf{Auto-S/L:} Auto-AVSR-Small/Large, \textbf{SL:} Supervised Learning, \textbf{SSL: }Self-Supervised Learning, \textbf{LLM:} Large Language Model Decoder, \textbf{PT: }Pre-training, \textbf{FT: }Fine-tuning, \textbf{LRS3: }WER in percentage (\%) as reported in original papers on LRS3 test set, \textbf{\bm{$\downarrow$}:} lower is better.}
\end{table}

We evaluate three representative pretrained VSR models spanning supervised, self-supervised, and large language model (LLM) paradigms. Auto-AVSR \cite{autoavsr} is a supervised conformer-based model with two variants, differing in training data scale (1,759 hours for Small and 3,291 hours for Large). AV-HuBERT \cite{avhubert} is a self-supervised model pretrained through masked multimodal prediction on 1,759 hours of unlabeled data, then finetuned on 433 hours of labeled data for recognition. VSP-LLM \cite{vspllm} integrates visual speech processing with large language model capabilities by encoding visual features through a pretrained AV-HuBERT frontend, and adapting them to a frozen Llama-2 \cite{llama2} backbone via Low-Rank Adaptation (LoRA) \cite{hu2022lora}. All models are evaluated using their publicly released pretrained weights, without additional finetuning. Table \ref{tab:models} summarizes the key characteristics of each model and includes performance on LRS3 \cite{lrs3} for reference.

\textbf{Video and Text Preprocessing:} Videos are processed using the standardized preprocessing pipeline from Auto-AVSR \cite{autoavsr} using RetinaFace \cite{RF}. Detected face regions are cropped and resized to 96×96 pixels, at 25 FPS, which serves as the standard input resolution for all VSR models. All text sequences are normalized by removing punctuation, converting to lowercase, and standardizing whitespace.


\subsection{Mapping the Mouth: Phonemes and Visemes}
\label{sec:mapping_mouth}

\begin{table}[t]
  \caption{Phoneme-to-Viseme mapping based on Microsoft Azure Speech Service \cite{msft_viseme_doc}. Each viseme groups phonemes that look visually similar during articulation.}
  \label{tab:viseme_mapping}
  \centering
  \footnotesize
  \begin{tabular}{@{}clll@{}}
    \toprule
    \textbf{ID} & \textbf{IPA} & \textbf{Example Words} & \textbf{Mouth Position} \\
    \midrule
    0  & Silence               & –                                  & Pause \\
    1  & \textipa{ae, @, 2}    & c\textbf{a}t, \textbf{a}bout, c\textbf{u}t & V: Open-mid \\
    2  & \textipa{A}           & f\textbf{a}ther, h\textbf{o}t            & V: Open-back \\
    3  & \textipa{O}           & c\textbf{augh}t, th\textbf{ough}t        & V: Open-mid back rounded \\
    4  & \textipa{E, U}        & b\textbf{e}d, p\textbf{u}t                & V: Open-front/ Close-back \\
    5  & \textipa{3`}          & b\textbf{ir}d, h\textbf{er}              & V: R-colored mid \\
    6  & \textipa{j, i, I}     & \textbf{y}es, b\textbf{ee}, b\textbf{i}t & V: High-front \\
    7  & \textipa{w, u}        & \textbf{w}e, b\textbf{oo}t               & V: High-back rounded \\
    8  & \textipa{o}           & b\textbf{oa}t, g\textbf{o}               & V: Mid-back rounded \\
    9  & \textipa{aU}          & c\textbf{ow}, \textbf{ou}t               & V: Diphthong \\
    10 & \textipa{OI}          & b\textbf{oy}, t\textbf{oy}               & V: Diphthong \\
    11 & \textipa{aI}          & b\textbf{uy}, \textbf{i}ce               & V: Diphthong \\
    12 & \textipa{h}           & \textbf{h}at, \textbf{h}ello             & C: Glottal fricative \\
    13 & \textipa{r}           & \textbf{r}ed, ca\textbf{r}               & C: Post-alveolar approximant \\
    14 & \textipa{l}           & \textbf{l}ip, ba\textbf{ll}              & C: Lateral approximant \\
    15 & \textipa{s, z}        & \textbf{s}it, \textbf{z}oo               & C: Alveolar fricative \\
    16 & \textipa{S, tS, dZ, Z}& \textbf{sh}e, \textbf{ch}in, \textbf{j}oy & C: Post-alveolar \\
    17 & \textipa{D}           & \textbf{th}is, brea\textbf{th}e          & C: Dental fricative \\
    18 & \textipa{f, v}        & \textbf{f}an, \textbf{v}an               & C: Labiodental fricative \\
    19 & \textipa{d, t, n, T}  & \textbf{d}og, \textbf{t}op, \textbf{n}o, \textbf{th}ink & C: Alveolar/dental \\
    20 & \textipa{k, g, N}     & \textbf{c}at, \textbf{g}o, si\textbf{ng} & C: Velar \\
    21 & \textipa{p, b, m}     & \textbf{p}at, \textbf{b}at, \textbf{m}at & C: Bilabial \\
    \bottomrule
  \end{tabular}
  \\[2pt]
  \footnotesize{Bold segments highlight the phonemes visually represented by each viseme, \textbf{ C:} Consonant, \textbf{V:}Vowel}
\end{table}

To enable analysis at both the phoneme and viseme levels (as detailed in later sections), we convert all model text predictions into phoneme sequences using the Phonemizer \cite{Bernard2021}. Phonemes are the smallest sound units distinguishing meaning, but many are visually indistinguishable on the lips (e.g., /p/, /b/, /m/ all show bilabial closure). Phonemes that are visually similar are mapped to viseme IDs according to the Microsoft Azure Speech Service 22-viseme classification (Table \ref{tab:viseme_mapping}), which is based on articulatory similarity \cite{msft_viseme_doc}. While other mappings also exist \cite{cappelletta2012phoneme,BEAR201740}, we use this one because its 22 classes capture subtle articulatory differences between visually similar phonemes, enabling a more detailed assessment of model viseme recognition. This enables direct comparison between model and human performance across phonemes and visemes. Since text normalization removes silence markers (viseme class 0), only viseme classes 1–21 are used in our analysis.

\section{Recognition Baselines: Humans vs Models}

We first establish baseline recognition performance across all VSR models on 2,189 words from the MaFI dataset \cite{krason2024mouth}. While word-level accuracy provides an overall performance measure, it does not indicate whether models are truly using visual cues, as high accuracy could result from reliance on linguistic regularities rather than visual articulation. To probe this, we also evaluate phoneme and viseme level performance. We evaluate performance using four metrics derived from the Levenshtein edit distance \cite{Levenshtein1965BinaryCC}, which counts the minimum number of insertions, deletions, and substitutions needed to transform one sequence into another. \textbf{Word error rate (WER)} and \textbf{character error rate (CER)} \cite{advocating_cer} measure transcription accuracy at word and character levels. \textbf{Phoneme score (PS)} and \textbf{viseme score (VS)} measure sequence similarity at phonetic and visemic levels. Text is converted to phoneme and viseme sequences using Phonemizer \cite{Bernard2021} and the Microsoft 22‑class mapping \cite{msft_viseme_doc} (See Section \ref{sec:mapping_mouth}). PS and VS reflect the proportion of correctly matched units between predicted and reference sequences based on standard edit-distance measures \cite{Levenshtein}, with higher values indicating better recognition. The human baseline performance for each metric (WER, CER, PS, VS) is computed by averaging the metric values across all participants for each of the 2,189 words. This allows for direct comparison between human and VSR model performance
.

\begin{table}
  \caption{Baselines on 2,189 words from MaFI dataset.}
  \label{tab:baseline}
  \centering
  \small
  \begin{tabular}{|l|c|c|c|c|}
    \hline
    \rowcolor{green!30!blue!20}
    \textbf{Model} & \textbf{WER \bm{$\downarrow$}} & \textbf{CER \bm{$\downarrow$}} & \textbf{PS \bm{$\uparrow$}} & \textbf{VS \bm{$\uparrow$}} \\
    \hline
    Human & 0.83 & 0.65  & 0.71 & 0.53 \\
    \hline
    Auto-AVSR-Large & \textbf{0.65} & \textbf{0.30}  & \textbf{0.87} & \textbf{0.82} \\
    \hline
    Auto-AVSR-Small & 1.11 & 0.56 & 0.75 & 0.66 \\
    \hline
    AV-HuBERT & 0.96 & 0.47 & 0.80 & 0.71  \\
    \hline
    VSP-LLM & 1.98 & 1.06  & 0.53 & 0.44\\
    \hline
  \end{tabular}
  \\[2pt]
  \footnotesize{\textbf{WER/CER:} Word/Character Error Rate, \textbf{VS/PS:} Viseme/Phoneme Score \textbf{\bm{$\downarrow$}:} lower is better, \textbf{\bm{$\uparrow$}:} higher is better, all metrics reported as decimals throughout this work (e.g., 0.83 WER = 83\%). }
\end{table}

Table \ref{tab:baseline} summarizes recognition performance across models and human participants. Auto-AVSR-Large achieves the best results (WER: 0.65, CER: 0.30), clearly outperforming human lipreaders (WER: 0.83, CER: 0.65). All models except VSP-LLM surpass human performance at both phoneme and viseme levels, despite identical visual input. Auto‑AVSR‑Large achieves PS 0.87 and VS 0.82, corresponding to 23\% and 55\% relative improvements over human scores (PS 0.71, VS 0.53). AV‑HuBERT and Auto‑AVSR‑Small show similar advantages, with Auto‑AVSR‑Small performing closer to humans on PS and VS while still achieving lower CER. These results align with the divergent confusion patterns discussed later in Figure \ref{fig:confusion_matrix}. Despite sharing the same visual constraints, most models exhibit higher viseme recognition accuracy than humans. Only VSP‑LLM underperforms humans across all metrics, reflecting its reliance on linguistic coherence rather than visual–articulatory cues. VSP-LLM's WER $>$ 1.0 reflects a tendency to generate multi-word outputs for single-word inputs.

\begin{table*}[ht]
  \caption{Model predictions and error metrics for representative examples with complete human guess lists.}
  \label{tab:prediction_examples}
  \centering
  \footnotesize
  \setlength{\tabcolsep}{4pt}
  \begin{tabular}{|>{\raggedright\arraybackslash}p{1.8cm}|>{\raggedright\arraybackslash}p{1.8cm}|>{\raggedright\arraybackslash}p{10.6cm}|c|c|}
    \hline
    \rowcolor{green!30!blue!20}
    \textbf{Ground Truth} & \textbf{Model} & \textbf{Prediction} & \textbf{WER \bm{$\downarrow$}} & \textbf{CER \bm{$\downarrow$}} \\
    \hline
    \textbf{acupuncture} & Auto-AVSR-L   & acupuncture         & 0.00 & 0.00 \\
    MaFI= -1.61                     & Auto-AVSR-S   & and compassion      & 2.00 & 1.09 \\
                         & AV-HuBERT     & anthropology        & 1.00 & 1.00 \\
                         & VSP-LLM       & neurobiology of     & 2.00 & 1.27 \\
    \hline
    \rowcolor{gray!10}
    \textbf{acupuncture} & Human (1/10)  & empathy, asteroid, horticulture, ammunition, accomplish, exercise, anaconda, immersion, brochure, acupuncture & 0.90 & 0.72 \\
    \hline
    \textbf{gondola}     & AV-HuBERT     & costs alive         & 2.00 & 1.29 \\
     MaFI= -1.91                    & Auto-AVSR-L   & yo silla            & 2.00 & 0.71 \\
                         & Auto-AVSR-S   & you are still alive & 4.00 & 2.29 \\
                         & VSP-LLM       & you are not alone   & 4.00 & 2.00 \\
    \hline
    \rowcolor{gray!10}
    \textbf{gondola}     & Human (0/10)  & half life, hotel, clockface, cook, cook out, nan, horseplay, love stare & 1.40 & 0.93 \\
    \hline
    \textbf{crosswalk}   & Auto-AVSR-L   & crosswalk           & 0.00 & 0.00 \\
    MaFI= -1.57                     & Auto-AVSR-S   & cross work          & 2.00 & 0.33 \\
                         & AV-HuBERT     & crosswalk           & 0.00 & 0.00 \\
                         & VSP-LLM       & crosswalk and       & 1.00 & 0.44 \\
    \hline
    \rowcolor{gray!10}
    \textbf{crosswalk}   & Human (3/21)  & crosswalk, brain, cosmo, cotswolds, hamburger, hashbrown, horeshoe, hospital, hotdogs, husband, iceberg, noisy, offspring, postmark, rocket, shakespeare, support, washer, world & 0.86 & 0.71 \\
    \hline
  \end{tabular}
  \\[2pt]
\footnotesize{
  \textbf{Auto-AVSR-L/S:} Auto-AVSR-Small/Large, \textbf{WER/CER:} Word/Character Error Rate,  \textbf{\bm{$\downarrow$}:} lower is better, 
  \textbf{Human (X/Y): }X correct out of Y total participants, 
  \textbf{MaFI = X:} MaFI score for that word.
}

\end{table*}

While baseline performance shows that most VSR models outperform humans, Table \ref{tab:prediction_examples} presents model and human predictions for words with complete human guess lists, illustrating how performance unfolds at the word and character levels. For a harder word with low MaFI (e.g., “gondola,” MaFI = -1.91), Auto-AVSR-Large achieves high accuracy, Auto-AVSR-Small produces a phonetically related alternative, AV-HuBERT predicts visually similar words, and VSP-LLM generates longer, context-driven phrases, whereas human guesses remain diverse, with all guesses incorrect. For an easier word (e.g., “crosswalk,” MaFI = -1.57), all models maintain near-perfect accuracy, while humans still produce multiple alternative guesses. These patterns indicate that models are not limited by visual information alone. Instead, they exploit patterns learned from training, such as frequent word sequences, recurring sound or syllable combinations, or context‑based predictions of likely phonemes. Human errors on the other hand, are influenced primarily by perceptual variability.

\section{Isolating Language Patterns from Visual Understanding}
\label{sec:lm_stuff}

While VSR models outperform humans (as seen in Section \ref{tab:baseline}), it remains unclear whether this stems from visual understanding or learned language patterns. VSR models are trained on large labeled corpora where they simultaneously learn visual-to-text mappings and language patterns. This raises two critical questions: First, how much recognition performance can be attributed to language patterns alone, independent of visual processing? Second, when VSR models make errors, are those errors driven by lack of visual information (like humans) or by gaps in their training vocabulary? To answer these questions, we first construct a text-only baseline with no visual input in Section 4.1, and then analyze whether model errors correlate more with training word frequency or with human visual difficulty in Section 4.2.

\subsection{Text-Only Language Model Baseline}
\label{sec:ngram}

To isolate the contribution of learned language patterns independent of visual processing, we constructed a text-only baseline that has no access to visual input. This baseline uses only the first K phonemes of each ground truth utterance as input. This minimal phonetic information simulates a scenario where the model relies purely on partial phonetic cues to predict the full word. If this text-only baseline with minimal context rivals human visual lipreading performance, it would demonstrate that language patterns are more powerful than visual information for word recognition. By evaluating on both MaFI (out-of-domain vocabulary) and LRS3 test (in-domain), we examine whether language patterns rely on memorizing training-specific vocabulary or generalize across domains.

We phonemized LRS3 training sentences using Phonemizer \cite{Bernard2021} (11.1M tokens, 49,928 unique words) and trained 2-gram and 5-gram models using KenLM \cite{heafield2011kenlm}, to learn phoneme sequence patterns from the training data, enabling word prediction from partial or complete phoneme sequences. For each test word, we extract the first K phonemes from the ground-truth word (e.g., K=3 from "absolutely" gives "æ b s") and retrieve all vocabulary words starting with these phonemes. The highest-scoring candidate is selected as the prediction, which is the sum of the n-gram language-model score for the candidate's full phoneme sequence and a log-transformed word-frequency ($\log_{10}(\text{word frequency} + 1)$). Accuracy is computed as the proportion of correct word matches.


\begin{table}
  \caption{Text-only n-gram baseline performance on MaFI (out-of-domain) and LRS3 (in-domain). The model receives only the first $K$ phonemes of each ground-truth word as input.}
  \label{tab:ngram_results}
  \centering
  \small
  \begin{tabular}{|l|c|c|c|c|}
    \hline
    \rowcolor{green!30!blue!20}
    \textbf{K} & \multicolumn{2}{c|}{\textbf{2-gram}} & \multicolumn{2}{c|}{\textbf{5-gram}} \\
    \hline
    \rowcolor{gray!15}
     & \textbf{MaFI} & \textbf{LRS3} & \textbf{MaFI} & \textbf{LRS3} \\
    \hline
    2    & 0.11 & 0.59 & 0.13 & 0.61 \\
    \hline
    3    & 0.41 & 0.77 & 0.41 & 0.79 \\
    \hline
    5    & 0.84 & 0.93 & 0.78 & 0.91 \\
    \hline
    Full & 0.94 & 0.98 & 0.86 & 0.95 \\
    \hline
  \end{tabular}
  \\[2pt]
  \footnotesize{$K$: number of initial phonemes provided as input. \textit{Full}: all phonemes are given to the model.}
\end{table}


Table \ref{tab:ngram_results} presents n-gram baseline results. With just K=2 phonemes and no visual input, the 2-gram achieves 11\% word accuracy on MaFI. With K=3, word accuracy rises to 41\%, surpassing human visual lipreading at 17\% (computed as 1 minus WER; see Table \ref{tab:baseline}) and approaching Auto-AVSR-Large at 35\% word accuracy (1-WER; see Table \ref{tab:baseline}). This demonstrates that much of the recognition challenge can be solved through linguistic patterns with minimal phoneme context alone, independent of visual processing.

The domain gap between MaFI (out-of-domain) and LRS3 (in-domain), shown in Table \ref{tab:ngram_results}, reveals the extent of this linguistic dependence. On the LRS3 test set, accuracy jumps to 59\% at K=2 and 77\% at K=3, which is 5× and 2× higher than MaFI for a 2-gram model. This gap shrinks as more phonemes are available (K=5: 84\% vs 93\%; full: 94\% vs 98\%), indicating that with partial phoneme information, the baseline exploits training vocabulary memorization. Since VSR models likely extract only partial phoneme-level features from visual input, their strong performance on LRS3 similarly stems from learned training patterns rather than robust visual understanding. Higher-order n-grams perform worse on MaFI (5-gram: 78\% vs 2-gram: 84\% at K=5), confirming overfitting to sentence-level patterns that fail to generalize beyond the training domain.

\subsection{Training Frequency vs Visual Difficulty}
\label{sec:freq_analysis}

To further determine whether VSR models rely primarily on memorized language patterns or visual speech processing, we analyze which factor better predicts model errors: training word frequency or human visual difficulty. Training frequency is defined as the log-transformed occurrence count of each word in the LRS3 training corpus, calculated as $\log_{10}(\text{count} + 1)$ (e.g., ``was'': 34,799 occurrences \(\rightarrow\) 4.54). Whereas visual difficulty is equivalent to MaFI scores. If models process visual speech similarly to humans, their WER should correlate more strongly with MaFI scores. Conversely, if models rely on memorized linguistic patterns, training frequency should be the stronger predictor. We therefore compute Spearman rank correlations between VSR model WER and (1) training word frequency and (2) MaFI scores for each evaluated model on MaFI words.

Across all models, training frequency exhibits consistently stronger correlations with WER than MaFI scores (mean \(|\rho|=0.35\) vs. 0.22, a 59\% relative difference). This pattern holds for Auto-AVSR-Large (0.35 vs. 0.24), Auto-AVSR-Small (0.37 vs. 0.26), AV-HuBERT (0.40 vs. 0.22), and VSP-LLM (0.28 vs. 0.17). These results indicate that high-frequency words are often predicted correctly despite visual ambiguity, whereas low-frequency words are more likely to fail even when visually distinct. Although the correlation magnitudes remain moderate $(|\rho| < 0.40)$, suggesting that model errors are influenced by multiple factors beyond either memorization or visual difficulty alone, the overall trend clearly shows that VSR models rely predominantly on memorized language patterns rather than human-like visual processing. We next examine whether models process visual information they receive in human-like ways.

\section{Reading the Lips: Viseme Ambiguity}
\label{sec:viseme}

\begin{figure*}[th]
  \centering
  \includegraphics[width=1\linewidth]{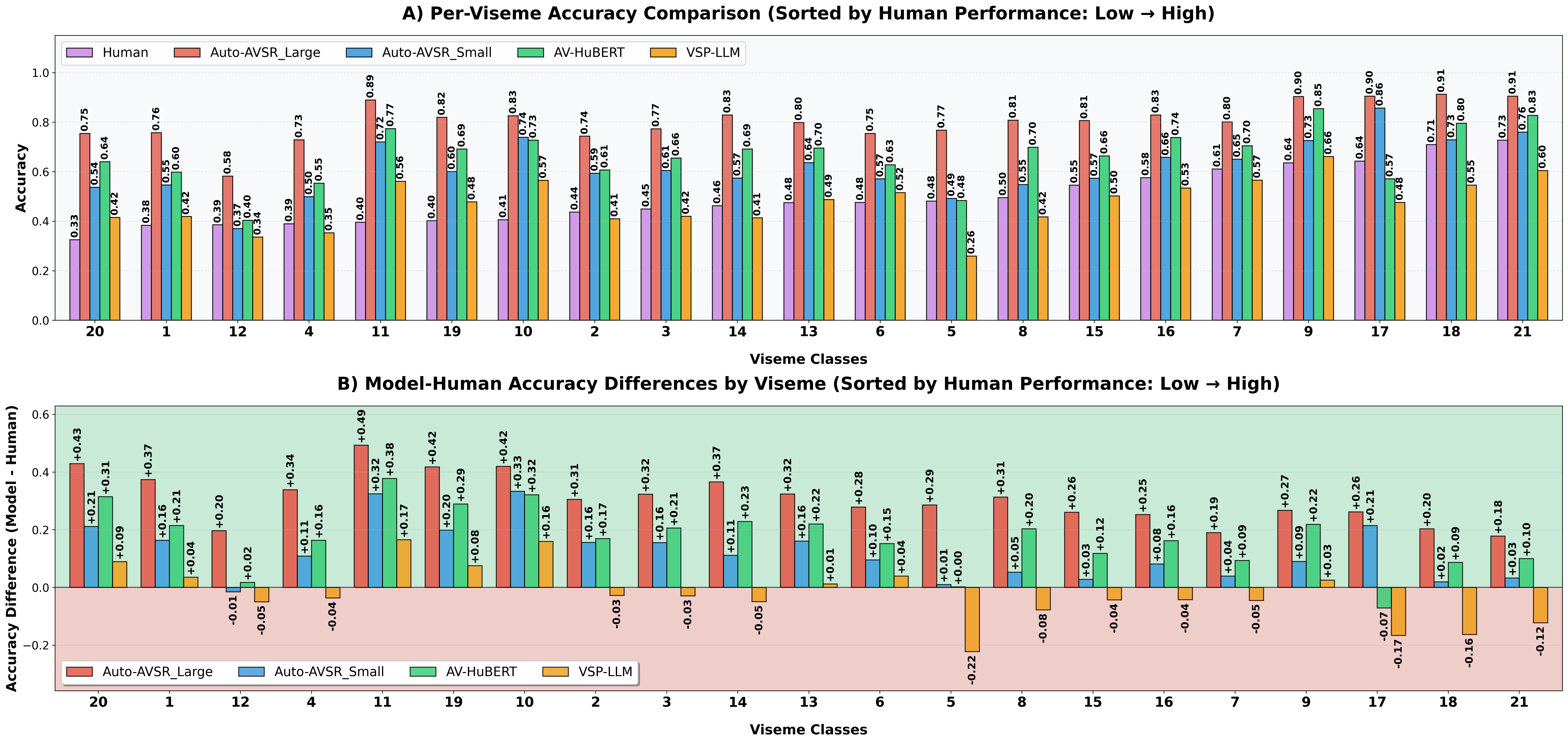}
  \caption{Viseme-level performance analysis: (A) Per-viseme accuracy comparison across models and humans, with visemes sorted by human performance. (B) Model-human accuracy differences where positive values indicate superior model performance. Viseme IDs in the figure correspond to the mapping presented in Table \ref{tab:viseme_mapping}.}
  \label{fig:viseme}
  \vspace{-10pt}
\end{figure*}

VSR is challenging because many phonemes appear visually similar or identical when spoken, a phenomenon known as viseme ambiguity \cite{BEAR201740}. Successful lipreading therefore relies on recognizing groups of phonemes that share common visual characteristics. To assess how VSR models handle this ambiguity compared to humans, we analyze viseme-level performance by first mapping text sequences to visemes (Section~\ref{sec:mapping_mouth}).

VSR models produce a single prediction per word. For each viseme class, VSR models’ per-class viseme accuracy is calculated as the number of correctly predicted visemes divided by the total number of times that viseme appears in the ground-truth sequences. In contrast, the MaFI dataset \cite{krason2024mouth} contains multiple guesses from different human participants for each word. Human per-class viseme accuracy is therefore calculated as the number of times participants correctly identified that viseme, divided by the total number of times the viseme appeared in the words presented to all participants. This calculation accounts for differences in viseme frequency and the fact that humans provide multiple guesses per word while models make a single prediction. Normalizing accuracy in this way ensures a fair, class-wise comparison of recognition performance between humans and VSR models.



\subsection{Class-Wise Lipreading Performance}

Figure \ref{fig:viseme}-A presents a per-viseme class accuracy comparison across humans and all VSR models. Visemes are presented by increasing human performance, from left to right.  A trend emerges whereby VSR models perform best on the exact viseme classes where humans perform worst. The five hardest visemes for humans (20: 33\%, 1: 38\%, 12: 39\%, 4: 39\%, 11: 40\%) are recognized by Auto-AVSR-Large at 75\%, 76\%, 58\%, 73\%, and 89\% respectively, showing 2-3x higher accuracy. These categories all lack clear visual cues and depend on subtle tongue movements, jaw configurations, or dynamic articulatory transitions that humans struggle to perceive, but models can detect through learned spatiotemporal patterns. In contrast, humans perform best on visemes with salient visible features (21: 73\%, 18: 71\%, and 17: 64\%) whereas model accuracies on these classes are also high but vary across systems, showing no consistent dependence on visual saliency. This divergent pattern suggests models are not just "better at lipreading" but actually use different information than humans, likely relying on statistical patterns learned from training data rather than individual mouth shapes. Auto-AVSR-Small and AV-HuBERT show similar performance patterns despite their different architectures and training methods, suggesting that training data volume primarily shapes both how models recognize visemes and the degree to which they align with human perception. VSP‑LLM diverged from this trend, performing below human level on several viseme classes and reflecting its architectural prioritization of contextual language modeling over visual feature extraction.

\subsection{Discrepancies in Visual Speech Processing: VSR Model VS Human Differences}

Figure \ref{fig:viseme}-B shows the per-viseme accuracy differences between VSR models and humans, highlighting where each model outperforms or underperforms compared to human performance. Auto-AVSR-Large outperforms humans on all viseme classes, with an average margin of 30.9 percentage points (pp) and the largest gains on visually ambiguous categories (e.g., viseme 11: +49.4pp, viseme 20: +42.9pp). Auto-AVSR-Small and AV-HuBERT follow the same trend with smaller average gains (+12.3pp and +17.1pp). An exception is viseme 17 (dental fricatives), where AV-HuBERT underperforms humans. In contrast, VSP-LLM shows near-human overall accuracy but actually performs worse than humans on visually clear classes, including r-colored vowels (–22.2pp), bilabials (–12.3pp), labiodentals (–16.3pp), and dentals (–16.6pp). This suggests that the model relies more on language‑based predictions than on visual information. The fact that all VSR models achieve disproportionately large gains on difficult visemes suggests that their performance relies on other features learned from training data rather than the visual cues humans use.

\subsection{Confusion Landscape of Lipreading}

\begin{figure}[ht]
  \centering
  \includegraphics[width=1\linewidth]{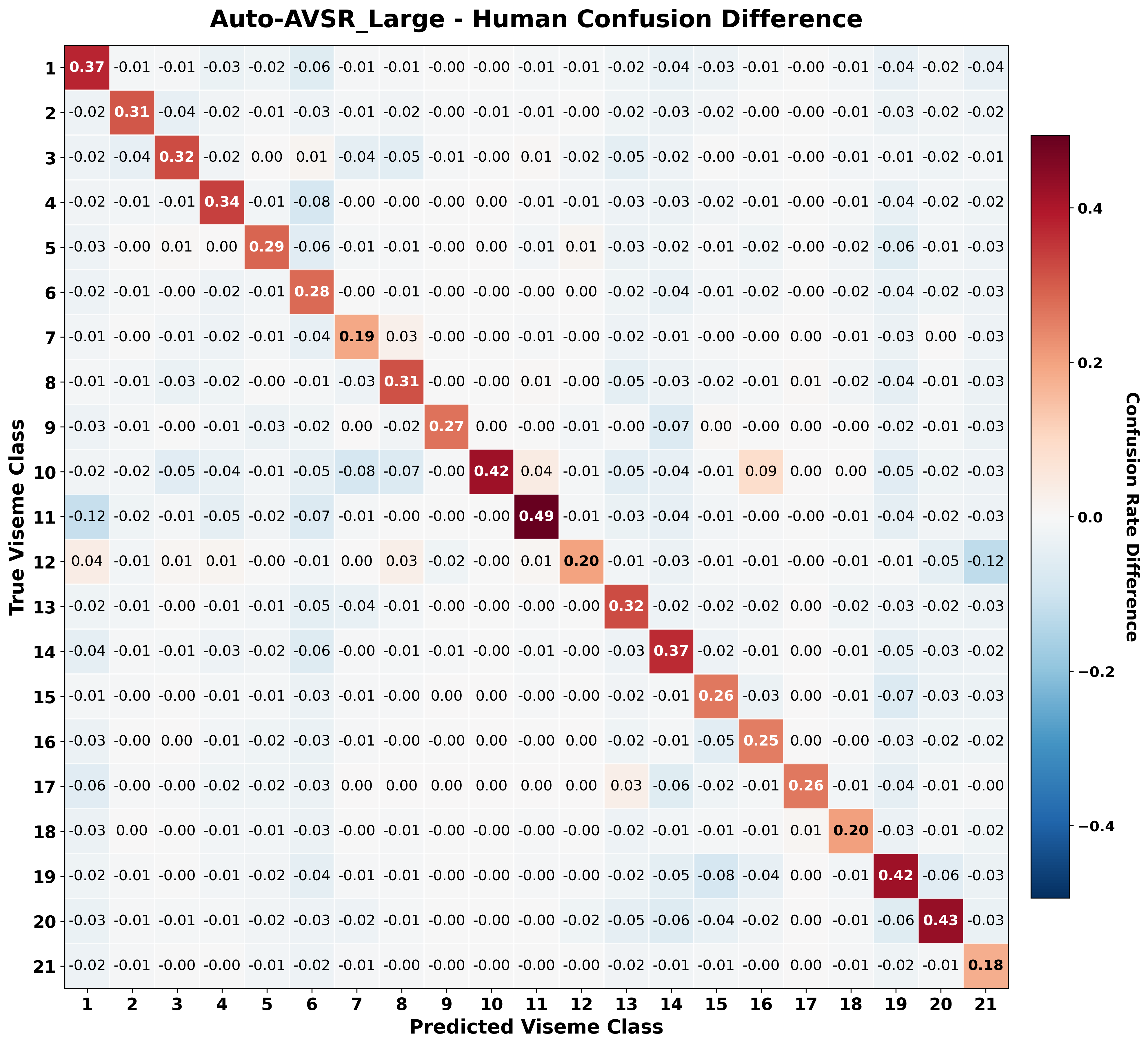}
  \caption{Viseme confusion differences between Auto‑AVSR‑Large and human baseline. Viseme IDs in the figure correspond to the mapping presented in Table \ref{tab:viseme_mapping}.}
  \label{fig:confusion_matrix}
  \vspace{-10pt}
\end{figure}

We also analyze model–human differences across viseme classes to determine whether recognition errors align with, or diverge from, human confusion patterns. Figure \ref{fig:confusion_matrix} presents a confusion difference matrix (Model minus Human) for Auto-AVSR Large. The model maintains strong positive diagonals, consistent with high recognition accuracy, but also displays systematic negative off‑diagonals where it makes substantially fewer confusions than humans  (12→21: –12.4pp, 11→1: –11.5pp, 10→7: –8.4pp, 19→15: –8.3pp, 4→6: –8.3pp). These negative values indicate viseme pairs that humans frequently confuse due to overlapping visual features, such as similar mouth shapes (4→6), lip rounding (10→7), or tongue positions (19→15), but VSR models rarely make these mistakes. This pattern suggests that the model disambiguates these cases using contextual cues rather than the visual similarity constraints that shape human perception. Auto‑AVSR‑Small and AV‑HuBERT exhibit nearly identical patterns, whereas VSP‑LLM shows a fundamentally different structure with language‑driven confusions (figures omitted due to space constraints). These patterns demonstrate that high accuracy does not guarantee human-like perception, though architectural choices can alter confusion topology. 

Across all viseme-level analyses, VSR models achieve higher accuracy than humans on visually ambiguous categories. Two explanations are possible: models either extract subtle visual temporal dynamics that humans cannot perceive, or compensate for visual ambiguity using learned linguistic patterns from their training data. Because errors correlate more strongly with training frequency than visual difficulty (as seen in \ref{sec:ngram}), the latter explanation is better supported.

\section{Human-Machine Alignment in Visual Speech Perception}
\label{sec:corr_align}

\begin{table*}[h]
  \caption{Correlation metrics and explained variance ($R^2$) across models and human performance.}
  \label{tab:correlation_metrics}
  \centering
  \small
  \begin{tabular}{|l|c|cccc|cccc|}
    \hline
    \rowcolor{green!30!blue!20}
    \textbf{Model/Human} & \textbf{HAS ($\rho$, $R^2$)} & \multicolumn{4}{c|}{\textbf{Human-Model Corr ($\rho$)}} & \multicolumn{4}{c|}{\textbf{MaFI Corr ($\rho$, $R^2$)}} \\
    \rowcolor{gray!15}
    \hline
    & & \textbf{WER} & \textbf{CER} & \textbf{VS} & \textbf{PS} & \textbf{WER} & \textbf{CER} & \textbf{VS} & \textbf{PS} \\
    \hline
    Human & (0.74, 0.55) & -- & -- & -- & -- & (-0.69, 0.53) & (-0.76, 0.50) & (0.88, 0.79) & (0.83, 0.70) \\
    \hline
    Auto-AVSR-Large & (0.37, 0.12) & 0.35 & 0.34 & 0.29 & 0.27 & (-0.24, 0.06) & (-0.26, 0.07) & (0.29, 0.09) & (0.27, 0.08) \\
    \hline
    Auto-AVSR-Small & (0.30, 0.10) & 0.28 & 0.34 & 0.31 & 0.26 & (-0.26, 0.07) & (-0.26, 0.06) & (0.30, 0.10) & (0.27, 0.08) \\
    \hline
    AV-HuBERT & (0.28, 0.08) & 0.26 & 0.32 & 0.31 & 0.27 & (-0.23, 0.05) & (-0.27, 0.06) & (0.28, 0.08) & (0.27, 0.07) \\
    \hline
    VSP-LLM & (0.17, 0.04) & 0.21 & 0.28 & 0.19 & 0.05 & (-0.18, 0.03) & (-0.14, 0.01) & (0.19, 0.04) & (0.10, 0.01) \\
    \hline
  \end{tabular}
  \\[2pt]
  \footnotesize{\textbf{HAS:} Spearman $\rho$ (human vs. model word accuracy). \textbf{Human-Model Corr:} Spearman $\rho$ (model vs. human per metric, per word). \textbf{MaFI Corr:} Spearman $\rho$ (metric performance vs. visual clarity). $R^2$ values computed as $\rho^2$, representing proportion of rank-order variance explained. $R^2$ values shown only for HAS and MaFI Correlation due to space constraints. All correlations significant at \textbf{$p<0.001$}.}
\end{table*}

Despite their overall strong performance on viseme accuracy, it remains unclear whether VSR models succeed and fail on the same words that humans find easy or difficult. To further examine this, we conducted correlation analyses \cite{kong2017evaluating} across multiple dimensions (Table \ref{tab:correlation_metrics}) to assess VSR models' alignment with human perceptual patterns. We first compute word-level alignment scores. To establish a human performance ceiling, we calculate \textbf{Human-Human Alignment Score (HHAS)} using split-half reliability testing \cite{eisinga2013reliability}: we randomly divided participant responses for each word into two groups over 100 iterations, computed Spearman correlations between group mean accuracies, and applied the Spearman-Brown correction (2ρ/(1+ρ)) \cite{eisinga2013reliability} to account for using half the data. This yields HHAS = 0.74 (Table \ref{tab:correlation_metrics}), representing the expected correlation between independent groups of human raters.  We then calculate \textbf{Human Alignment Score (HAS)} \cite{han2025video} for each model, defined as the Spearman rank correlation between human word accuracy (proportion of participants correct per word) and model word accuracy (binary correct/incorrect per word). Beyond word-level alignment, we measure per-metric correlations between human and model performance for each metric (WER, CER, VS, and PS) to evaluate whether models struggle on the same words as humans. We also quantify dependence on visual information through MaFI correlations, defined as the Spearman correlation between MaFI scores and performance metrics for both VSR models and humans.

\textbf{Human Alignment Scores (left column)} reveal weak word-level agreement across all models. Auto-AVSR-Large achieves the highest HAS (ρ = 0.37, R² = 0.12), meaning only 12\% of rank-order variance in model word-level accuracy patterns is explained by human patterns. Auto-AVSR-Small and AV-HuBERT show similar weak alignment (0.30 and 0.28 respectively), while VSP-LLM exhibits the weakest correlation (ρ = 0.17, R² = 0.04). These low HAS values demonstrate that models do not reliably succeed on the same words humans find easy, showing only weak agreement on word-level difficulty.  Notably, all model HAS values are substantially below the human performance ceiling (HHAS = 0.74), indicating that even the best-performing model captures only about half of the word-level rank-order consistency seen among human participants. 

\textbf{Human-model correlations (middle columns)} show modest overlap in error patterns ranging from ρ=0.05 to 0.35. Auto-AVSR-Large averages 0.31 across metrics, suggesting partial alignment but broad disagreement on word difficulty. Most VSR models achieve their strongest correlations with human CER (ρ = 0.28-0.34), indicating better alignment at character-level errors than complete word recognition. VSP-LLM shows dramatically weaker PS correlation (ρ = 0.05) compared to other VSR models (ρ = 0.26-0.27), suggesting its language model component generates phoneme-level predictions that diverge sharply from human perceptual patterns.

\textbf{MaFI correlations (right columns)} reveal the most critical difference. Humans show strong correlations across all metrics (VS: ρ=0.88, PS: ρ=0.83, CER: ρ=-0.76, WER: ρ=-0.69), with R² values ranging from 0.50 to 0.79, demonstrating that visual clarity explains 50-79\% of human performance variance. The negative values for error rates indicate that as visual clarity increases, errors decrease. This cross-level consistency demonstrates humans extract more information when visual cues are clear and recognize perceptual limits when cues are ambiguous. In contrast, VSR models show substantially weaker correlations with MaFI scores. Auto-AVSR-Large achieves ρ=0.29 for VS and ρ=0.27 for PS (3x weaker than humans), with error rate correlations of only -0.24 and -0.26 (explaining just 6-7\% variance versus humans' 50-53\%). VSP-LLM shows even weaker correlations (VS: ρ = 0.19, PS: ρ = 0.10, R² = 0.01-0.04), consistent with its reliance on linguistic priors rather than visual features. Overall, current VSR models diverge substantially from human perceptual processing, showing weaker correlations to visual clarity and weaker alignment with human error patterns. One might argue that the strong human-MaFI correlation is circular, since MaFI derives from human responses. However, MaFI quantifies visual properties of words (via phonological edit distance), not individual accuracy. The strong human correlation demonstrates systematic constraint by visual properties, while the 3× weaker model correlation (ρ=0.27-0.29 vs. ρ=0.83-0.88) reveals models are less constrained by visual clarity.

\begin{table*}[h]
  \caption{Performance split by visual clarity and percentage decline from high to low clarity.}
  \label{tab:clarity_decline_full}
  \centering
  \small
  \begin{tabular}{|l|cccc|cccc|cccc|}
    \hline
    \rowcolor{green!30!blue!20}
    \textbf{Model/Human} & \multicolumn{4}{c|}{\textbf{High Clarity (MaFI $>$ -1)}} & \multicolumn{4}{c|}{\textbf{Low Clarity (MaFI $\leq$ -1)}} & \multicolumn{4}{c|}{\textbf{Relative Difference (\%)}} \\
    \rowcolor{gray!15}
        \hline
    & \textbf{WER $\downarrow$} & \textbf{CER $\downarrow$} & \textbf{VS $\uparrow$} & \textbf{PS $\uparrow$} & \textbf{WER $\downarrow$} & \textbf{CER $\downarrow$} & \textbf{VS $\uparrow$} & \textbf{PS $\uparrow$} & \textbf{WER} & \textbf{CER} & \textbf{VS} & \textbf{PS} \\
    \hline
    Human & 0.71 & 0.51 & 0.66 & 0.79 & 0.95 & 0.79 & 0.41 & 0.63 & +34 & +55 & -38 & -20 \\
    \hline
    Auto-AVSR-Large & 0.48 & 0.24 & 0.88 & 0.91 & 0.79 & 0.41 & 0.76 & 0.84 & +64 & +71 & -14 & -8 \\
    \hline
    Auto-AVSR-Small & 0.88 & 0.48 & 0.74 & 0.80 & 1.30 & 0.72 & 0.59 & 0.71 & +48 & +50 & -20 & -11 \\
    \hline
    AV-HuBERT & 0.78 & 0.39 & 0.78 & 0.85 & 1.11 & 0.59 & 0.65 & 0.77 & +42 & +52 & -17 & -9 \\
    \hline
    VSP-LLM & 1.81 & 1.10 & 0.49 & 0.56 & 2.10 & 1.22 & 0.41 & 0.52 & +16 & +11 & -16 & -7 \\
    \hline
  \end{tabular}
  \\[2pt]
\footnotesize{
\textbf{WER/CER:} Word/Character Error Rate, \bm{$\downarrow$}: lower is better; \textbf{VS/PS:} Viseme/Phoneme Scores,  $\bm{\uparrow}$: higher is better
; \textbf{Relative Difference:} percentage change from high to low clarity \((\text{low} - \text{high}) / \text{high} \times 100\);
\textbf{High Clarity:} \(N=1056\) Easy words; \textbf{Low Clarity:} \(N=1133\) Hard words.
}
\end{table*}

\section{Performance by Visual Information Clarity}
\label{sec:mafi_divide}

\begin{figure}[ht]
  \centering
  \includegraphics[width=1\linewidth]{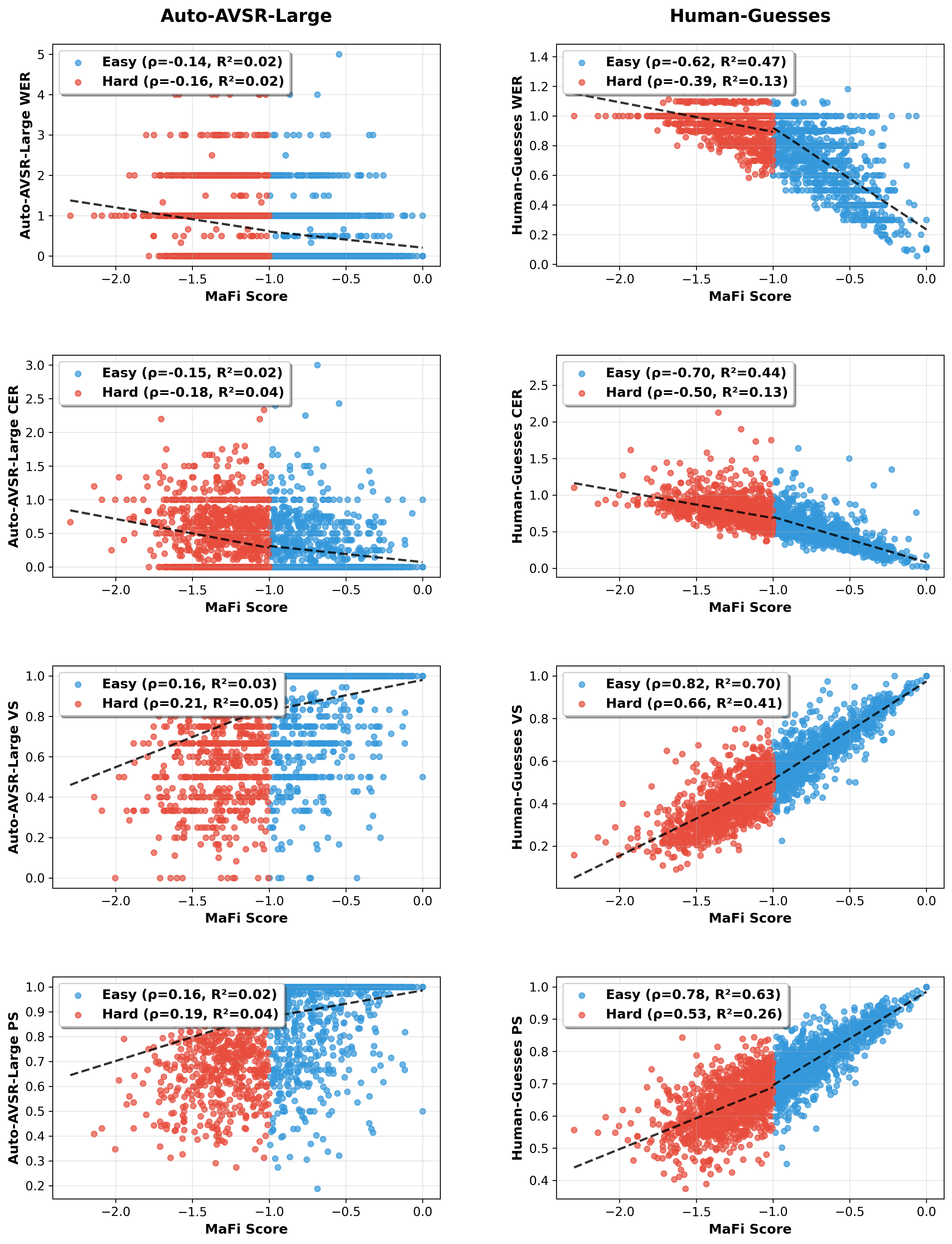}
  \caption{Spearman correlations between MaFI scores and performance metrics for Auto-AVSR-Large and Human Guesses on easy (high-clarity, $MaFI > -1$) and hard (low-clarity, $MaFI \leq -1$) words.}
  \label{fig:clarity}
  \vspace{-10pt}
\end{figure}

To further assess how the visual saliency of words affects recognition performance, we group them according to their MaFI score, reflecting visual clarity. We split the dataset into high-clarity (easy) words ($MaFI > -1$, $N=1056$, visually clear) and low-clarity (hard) words ($MaFI \leq -1$, $N=1133$, visually ambiguous). For each clarity level, we compute WER, CER, VS, and PS, measuring performance changes from high to low clarity conditions (Table \ref{tab:clarity_decline_full}). We also compute Spearman correlations between MaFI Score and performance metrics within each clarity level. Figure \ref{fig:clarity} presents these correlations between MaFI scores and performance metrics for Auto-AVSR-Large and Human responses across easy versus hard words. Due to space constraints, we do not provide plots for the remaining models, though we detail their correlation patterns later in this section. 

Table \ref{tab:clarity_decline_full} reveals a  mismatch between sub-lexical (VS/PS) and lexical-level (WER/CER) errors as visual clarity declines. Humans show large drops in sub-lexical scores (VS: $-38\%$, PS: $-20\%$) but moderate increases in transcription errors (WER: +34\%, CER: +55\%). VSR models show the opposite: Auto-AVSR-Large maintains strong articulation scores (VS: -14\%, PS: -8\%) but shows much worse recognition accuracy (WER: +64\%, CER: +71\%). Other models fall between these extremes: Auto-AVSR-Small and AV-HuBERT show VS drops of -20\% and -17\% and PS drops of -11\% and -9\%, with error increases of +42\% to +52\%. VSP-LLM exhibits the smallest changes across all metrics, with VS and PS declining by -16\% and -7\% and lexical errors increasing by only +11\% and +16\%.

Comparing Auto-AVSR-Large and Human Guesses (Figure \ref{fig:clarity}) reveals opposing correlation patterns that explain this mismatch in Table \ref{tab:clarity_decline_full} . Humans show stronger correlations on easy words (VS: $\rho$=0.82, PS: $\rho$=0.78) than hard words (VS: $\rho$=0.66, PS: $\rho$=0.53), performing better on visually clear input. Auto-AVSR-Large exhibits the reverse: stronger correlations on hard words across all metrics (VS: $\rho$= 0.21 vs 0.16, PS: $\rho$= 0.19 vs 0.16, CER: $\rho$= -0.18 vs -0.15, WER: $\rho$= -0.16 vs -0.14). This pattern occurs because Auto-AVSR-Large's sub-lexical accuracy decreases only 14\% for visemes and 8\% for phonemes (VS: 0.88→0.76, PS: 0.91→0.84) from easy to hard words, whereas humans drop 38\% and 20\% (VS: 0.66→0.41, PS: 0.79→0.63). These smaller declines allow Auto-AVSR-Large to maintain distinctions between difficulty levels on hard words, while humans' larger drops reduce their ability to make such distinctions. Despite this difference, Auto-AVSR-Large fails to translate it into accurate word-level transcription, resulting in larger WER and CER increases than humans. Notably, both humans and models show larger relative increases in CER than WER, indicating that visual ambiguity primarily affects character-level segmentation rather than whole-word confusions. 

Model-specific variations reveal systematic differences in correlation patterns. VSR models trained on larger datasets show greater divergence from human visual speech perception. Auto-AVSR-Small exhibits mixed patterns: stronger VS/PS correlations on hard words but stronger WER/CER correlations on easy words (more human-like). AV-HuBERT shows even more human-like error patterns but differs from humans at sub-lexical level. VSP-LLM displays human-like patterns across all metrics but with weak overall correlations ($\rho$ $\sim$ 0.08-0.12). This progression suggests that scaling training data improves recognition accuracy but leads models to develop perceptual strategies that increasingly diverge from human perceptual mechanisms. Notably, even the strongest model correlations (VS: $\rho$=0.21) remain approximately 3$\times$ weaker than humans' on hard words (VS: $\rho$=0.66), as shown in Figure \ref{fig:clarity}. This reveals fundamentally different processing mechanisms despite superior absolute performance (Auto-AVSR-Large VS: 0.76 vs human VS: 0.41 on hard words, Table \ref{tab:clarity_decline_full}). Superior model performance does not reflect human-like visual speech processing, highlighting the need for evaluation approaches that distinguish perceptual mechanisms from statistical learning.

\section{Discussion and Conclusion}

VSR research implicitly equates improvements in transcription accuracy (such as WER) with progress in visual speech understanding. Our findings suggest this assumption is incomplete. Given the first three phonemes as input, text-only n-grams achieve 76.7\% accuracy on in-domain LRS3 vocabulary but only 41\% on MaFI, revealing substantial reliance on learned linguistic patterns and overfitting to an in-domain dataset. The best-performing model Auto-AVSR-Large achieves 0.65 WER on the MaFI Dataset (0.20 WER on LRS3) and yet it shows weak human alignment (HAS: 0.37, 12\% variance) compared to the human ceiling (HHAS: 0.74). This begs the question of whether VSR models are even intended to process visual information like humans do. One might argue it does not matter: if models produce low WER, why care how they do it? Yet how models achieve that accuracy reveals a critical vulnerability. Our findings showed that VSR models maintain high viseme accuracy while word errors increase 64\% on ambiguous words, showing they perceive lip movements but cannot map them to correct words when visual clarity drops. Models also excel on visually ambiguous categories where humans struggle, making fewer confusions on difficult visemes, but missing the visual-perceptual patterns that shape human errors. Humans rely on visual articulation cues alongside linguistic prediction to resolve ambiguity \cite{peelle2015prediction}, whereas VSR models depend almost entirely on learned linguistic patterns, making them vulnerable when deployed on new speakers and unfamiliar visual input. This is further evidenced by VSR model WER correlating more strongly with training frequency (\(|\rho|=0.35\)) than visual difficulty (\(|\rho|=0.22\)). Notably, recent trends toward LLM-focused VSR \cite{vspllm, cappellazzo2025large, Yeo_2025_ICCV}, aimed at improving contextual understanding do not mitigate this issue: the VSP-LLM model exhibits the largest errors on unseen MaFI data, producing multiple words for single-word inputs and highlighting a limitation in capturing visual articulation. Recent work also shows that LLM‑based VSR gains stem primarily from contextual or lexical reasoning, not improved visual integration, reinforcing our findings \cite{jain2026hypeinsightrethinkinglarge}.

These findings have practical implications for VSR deployment. In audio-visual scenarios where visual and acoustic information conflict, such as cocktail-party situations \cite{nguyen25b_interspeech} or McGurk effect conditions \cite{mcgurk1976hearing}, systems that depend on linguistic prediction rather than visual articulation may fail unpredictably. Current evaluation using only WER cannot distinguish systems that rely on visual articulation cues from those that rely on learned language patterns. Moreover, weaker MaFI correlations in models (ρ=0.29 for VS, 3x weaker than humans' ρ=0.88) indicate that models do not adapt their performance to visual clarity in the same manner as humans. Architectural diversity alone, as demonstrated across supervised, self-supervised, and LLM-based approaches, cannot overcome limitations of standard cross-entropy and CTC loss functions. These loss functions reward any signal that minimizes prediction error, encouraging linguistic shortcuts that bypass visual articulation cues. Addressing this fundamental mismatch requires rethinking training objectives to explicitly reward alignment with visual articulation rather than just error minimization. 

A limitation of this work is that we only evaluate on isolated English words from the MaFI dataset, without examining continuous speech or extending to other languages. Such an extension would require comparable publicly-available human lipreading experiments, which do not currently exist to the best of our knowledge.

Advancing VSR requires moving beyond benchmark scores towards accurate modeling of visual articulatory cues and human-aligned perception. Future architectures should explore explicit visual attention mechanisms \cite{dosovitskiy2021image} that focus on visually salient facial regions, perceptual loss functions \cite{BRUCKERT2021693} that encourage matching human-like difficulty patterns \cite{wang2024predict,liu2025human,zhang2020can}, or Vision Transformer-based encoders \cite{arnab2021vivit,dosovitskiy2021image} whose attention patterns may better align with human visual processing  \cite{Thomas_2025_ICCV, park2025swinlip}. Multi-task learning could jointly optimize both lexical (word/character) and sub-lexical (phoneme/viseme-level) accuracy \cite{kim2024efficient, kit2025phoneme}, while extending this analysis to continuous speech and multiple languages would strengthen generalizability \cite{kim2024efficient}.  Evaluation frameworks must move beyond transcription accuracy to incorporate human-perception alignment and multi-level phoneme/viseme metrics \cite{wada2024polos,demircan2024evaluating, kong2017evaluating}. Building on these findings, we plan to develop evaluation metrics that integrate both phonetic and visemic information with human alignment measures, and to incorporate phoneme- and viseme-based loss objectives into AVSR training \cite{papadopoulos2025interpreting, kit2025phoneme}. This shift toward human-centered evaluation will be essential for creating VSR systems that are robust, interpretable, and aligned with human perceptual mechanisms \cite{liu2025human, Yuan_2025_ICCV}.

\section{Acknowledgments}
This publication emanates from research supported by Taighde Éireann – Research Ireland, Grant number 22/FFP-A/11059.

\section{Generative AI Use Disclosure}
During the preparation of this work, Claude (Anthropic) was used only for minor English grammar corrections and refining the clarity of written content.

\bibliographystyle{IEEEtran}
\bibliography{mybib}

\end{document}